\documentclass[sigconf]{acmart}

\AtBeginDocument{
  }

\setcopyright{none}
\acmConference{CIKM 2025 FinAI Workshop}
\acmBooktitle{}
\acmISBN{}
\acmDOI{}
\acmYear{2025}
\copyrightyear{}
\date{}

\renewcommand\footnotetextcopyrightpermission[1]{}

\usepackage[most]{tcolorbox}
\tcbuselibrary{listings,skins,breakable}
\usepackage{tabularx}
\usepackage{float}
\usepackage{longtable}

\usepackage{enumitem}

\begin{document}

\title{Interpreting LLMs as Credit Risk Classifiers: Do Their Feature Explanations Align with Classical ML?}


\newcommand{\ADIA}{%
  \affiliation{%
    \institution{ADIA}
    \city{Abu Dhabi}
    \country{United Arab Emirates}
  }%
}
\newcommand{\KU}{%
  \affiliation{%
    \institution{Khalifa University}
    \city{Abu Dhabi}
    \country{United Arab Emirates}
  }%
}

\author{Saeed AlMarri}
\KU
\email{100061460@ku.ac.ae}

\author{Kristof Juhasz}
\ADIA
\email{kristof.juhasz@adia.ae}

\author{Mathieu Ravaut}
\ADIA
\email{mathieu.ravaut@adia.ae}

\author{Gautier Marti}
\ADIA
\email{gautier.marti@adia.ae}

\author{Hamdan Al Ahbabi}
\KU
\email{100061346@ku.ac.ae}

\author{Ibrahim Elfadel}
\KU
\email{ibrahim.elfadel@ku.ac.ae}

% Running head (replaces \authorrunning from LNCS)
\renewcommand{\shortauthors}{AlMarri et al.}


\begin{abstract}
  Large Language Models (LLMs) are increasingly explored as flexible alternatives to classical machine learning models for classification tasks through zero-shot prompting. However, their suitability for structured tabular data remains underexplored, especially in high-stakes financial applications such as financial risk assessment. This study conducts a systematic comparison between zero-shot LLM-based classifiers and LightGBM, a state-of-the-art gradient-boosting model, on a real-world loan default prediction task. We evaluate their predictive performance, analyze feature attributions using SHAP, and assess the reliability of LLM-generated self-explanations. While LLMs are able to identify key financial risk indicators, their feature importance rankings diverge notably from LightGBM, and their self-explanations often fail to align with empirical SHAP attributions. These findings highlight the limitations of LLMs as standalone models for structured financial risk prediction and raise concerns about the trustworthiness of their self-generated explanations. Our results underscore the need for explainability audits, baseline comparisons with interpretable models, and human-in-the-loop oversight when deploying LLMs in risk-sensitive financial environments.
\end{abstract}






\maketitle

\section{Introduction}

Large Language Models (LLMs), such as GPT-4, have demonstrated strong performance across a range of natural language processing (NLP) tasks, including classification and reasoning \cite{vaswani2017attention,brown2020language,achiam2023gpt}. Their ability to function as classifiers without explicit training pipelines, relying solely on a few-shot or zero-shot prompting, has gained significant attention. This raises fundamental questions about the reliability and validity of LLM-based classification, particularly in comparison to classical machine learning models such as gradient-boosting decision trees methods like XGBoost \cite{chen2016xgboost} or LightGBM \cite{ke2017lightgbm}.

Traditional classification tasks require structured pipelines involving feature engineering, model training, validation, and hyperparameter tuning. Fine-tuning models on tabular data, in particular, demands expertise in preprocessing, GPU management, and balancing class distributions to prevent trivial solutions. In contrast, LLMs bypass fine-tuning entirely, requiring only natural language prompting. This reduces technical barriers, making them accessible to non-experts, but raises an important question: \textit{How do LLM-based classifiers arrive at their predictions, and do they rely on decision patterns similar to classical machine learning models?}

This question is especially critical in high-stakes financial domains, where algorithmic risk assessments directly affect credit access, interest rates, and regulatory compliance \cite{doshi-velez2017towards}. Financial institutions operate under strict governance frameworks such as Basel III \cite{BaselIII} and  GDPR \cite{GDPR}, where opaque models can lead to regulatory breaches, reputational damage, and unfair or discriminatory decisions. Unlike decision trees or gradient boosting models, LLMs are complex black-box models with billions of parameters, making interpretability a key challenge. This has led to growing interest in Explainable AI (XAI) techniques to analyze LLMs’ internal logic and assess their alignment with human-interpretable decision patterns.

In this study, we conduct a systematic evaluation of zero-shot LLM classifiers for structured credit risk prediction, directly comparing them with a well-established interpretable baseline (LightGBM). Beyond performance comparison, our primary goal is to audit their explainability to determine whether their feature attributions and self-generated rationales align with dataset-driven reasoning or rely on external priors. We employ Shapley Additive Explanations (SHAP) \cite{lundberg2017} to analyze the faithfulness of their decision patterns. Leveraging a public loan default prediction dataset, we address the following research questions:

\begin{itemize}[leftmargin=*, rightmargin=0pt]
    \item \textbf{Explainability:} Do LLM-based classifiers prioritize the same features as classical models?
    
    \item \textbf{Self-Explainability:} Can LLMs provide self-generated rationales that align with SHAP-derived feature attributions?
    
    \item \textbf{Performance:} How do LLMs compare to classical machine learning classifiers (ROC-AUC, PR-AUC)? Does ensembling both types of models improve performance?
\end{itemize}

Our work directly addresses the workshop call on AI safety, fairness, and explainability in high-stakes financial environments, and responsible deployment in fintech and banking, by auditing the faithfulness of LLM explanations against SHAP on a real credit-risk task. The remainder of this paper is structured as follows: Section 2 reviews related work, Section 3 details the methodology, Section 4 presents the experimental setup, Section 5 analyzes results, feature attribution and reasoning patterns, and Section 6 concludes with key findings and future directions.

\section{Related Work}

%%%%%%%%%%%%%%%%%%%%%%%%%%%%%%%%%%%%%%%%%%%%%%%%%%%%
\subsection{LLMs for Zero-Shot Classification}

LLMs have demonstrated strong classification capabilities, often achieving competitive performance without supervision. 
\citet{brown2020language} introduced GPT-3, highlighting its few-shot and zero-shot classification potential. 
This paradigm removes the need for classical training pipelines, enabling non-experts to perform classification via direct natural-language prompting and in-context learning. 
Subsequent work has explored structured prompting to enhance classification accuracy. 
\citet{hao2020selfattention} introduced Chain-of-Thought (CoT) prompting, showing that structured reasoning can improve LLM performance—relevant when adapting them to structured data. 

A recent line of work examines whether LLMs can function as regressors for numerical data. 
\citet{vacareanu2024from} found that LLMs approximate regression functions with in-context examples, while \citet{buckmann2024logistic} proposed combining LLMs with logistic regression for low-data robustness. 
\citet{song2024omnipred} and \citet{song2025decoding} further extend this research direction, demonstrating universal regression capabilities using decoding-based inference.

\paragraph{\textbf{Our Contribution}} Unlike prior studies that evaluate LLM classification in isolation, our work positions this comparison as a step toward responsible AI deployment in high-stakes financial contexts. By systematically comparing zero-shot LLM and LightGBM on the same dataset and auditing their decision patterns using SHAP-based explainability, we assess not only accuracy but also faithfulness and reliability of model reasoning critical elements for trustworthy, fair and transparent AI use in credit risk assessment.

%%%%%%%%%%%%%%%%%%%%%%%%%%%%%%%%%%%%%%%%%%%%%%%%%%%%
\subsection{LLM Explainability}

Feature-attribution methods such as SHAP \cite{lundberg2017} are widely used to assess feature importance in machine-learning models. 
We use SHAP to compare LLM-based probabilistic classifiers with LightGBM. 
A key question is whether LLMs’ self-explanations align with actual feature importance. 
\citet{huang2023can} report that LLM rationales are often plausible but do not necessarily reflect internal reasoning. 
\citet{dehghanighobadi2025can} analyze counterfactual explanations and show that LLMs can struggle with causal dependencies. 
\citet{sarkar2024large} argues that LLMs lack self-explanatory capabilities due to opaque training dynamics, and \citet{turpin2023language} show that CoT-generated explanations can be misleading.

\paragraph{\textbf{Our Contribution}} Prior work predominantly studies LLM self-explanations in text tasks. To our knowledge, our study is the first to compute SHAP-based feature importance for LLMs prompted to output probabilistic predictions on structured financial data, enabling a direct faithfulness audit of self-explanations against empirical attributions.

%%%%%%%%%%%%%%%%%%%%%%%%%%%%%%%%%%%%%%%%%%%%%%%%%%%%
\subsection{LLMs for Tabular Data}

Recent work explores whether LLMs can replace gradient-boosted models such as XGBoost, LightGBM, and AdaBoost for tabular classification.
\citet{fang2024large} survey LLMs on tabular data and highlight adaptation challenges. 
\citet{ghaffarzadeh2024large} benchmark LLMs against classical ML for COVID-19 mortality prediction, concluding that classical ML models outperform LLMs on structured data.
\citet{chen2024clinicalbench} introduce ClinicalBench and similarly find XGBoost superior for clinical prediction tasks. 
\citet{hegselmann2023} propose TabLLM, which reformulates tables as natural language for few-shot classification, while \citet{shi2024} introduce ZET-LLM, treating autoregressive LLMs as feature-embedding models for tabular prediction. 
While these studies highlight LLM potential, they generally do not evaluate explainability or faithfulness of rationales on tabular tasks.

\paragraph{\textbf{Our Contribution}} We conduct a head-to-head comparison of LLMs and LightGBM on the same structured dataset and integrate SHAP-based explainability, offering a dual analysis of predictive performance and feature attribution to illuminate decision mechanisms.

%%%%%%%%%%%%%%%%%%%%%%%%%%%%%%%%%%%%%%%%%%%%%%%%%%%%
\subsection{Classical ML, XAI, and LLMs for Financial AI}

Explainability is crucial in financial applications for risk assessment. 
\citet{martins2023explainable} review XAI in finance, while \citet{vcernevivciene2024explainable},  \citet{misheva2021explainable}, and  \citet{bussmann2021explainable} analyze explainability in credit-risk modeling. 
Several studies benchmark ML models for loan-default prediction: \citet{madaan2021loan} compare decision trees and random forests without an explainability analysis; \citet{srinivasa2021comparative} assess ML techniques for loan risk but do not explore LLMs; and \citet{boughaci2020appropriate} study credit-scoring models without evaluating LLM-based predictions.

\paragraph{\textbf{Our Contribution}} While prior work focuses on classical ML for loan prediction, we present the first \textit{comparative analysis of LLMs and LightGBM on structured loan data, integrating SHAP-based explainability}. Our findings extend beyond credit risk to broader financial applications, including fraud detection, regulatory compliance, and algorithmic trading decision-making, where explainability is key to adoption.

% \subsection{Synthesis and Responsible-AI Gap}
% Most tabular–LLM studies prioritize accuracy and prompting but rarely test whether the LLM rationales are \emph{faithful}, \emph{auditable}, or deployment-ready in credit settings. They rarely (i) validate self-explanations against model-agnostic attributions (e.g. SHAP), (ii) situate results within model risk management (MRM) practice, or (iii) examine fairness and compliance implications despite credit scoring being regulated and, in the EU, classified as high-risk AI with duties for transparency and human oversight \cite{EUAIAct_2024, EBA_2020_LoanOrigination}.

% \textbf{Our Contribution.} We use SHAP as a model-agnostic \emph{audit} baseline to benchmark LLM self-rationales against empirical attributions on loan-default prediction. The protocol mirrors banking MRM (robust development/use, independent validation, ongoing monitoring) \cite{frb_sr11_7_2011} and aligns with AI risk guidance \cite{NIST_2023_AI_RMF}. SHAP is not treated as causal ground truth, but as a consistent reference for auditing rationale–decision alignment to inform go/no-go and monitoring design.

\section{Methodology}

\begin{figure}
    \centering
    \includegraphics[width=1\linewidth]{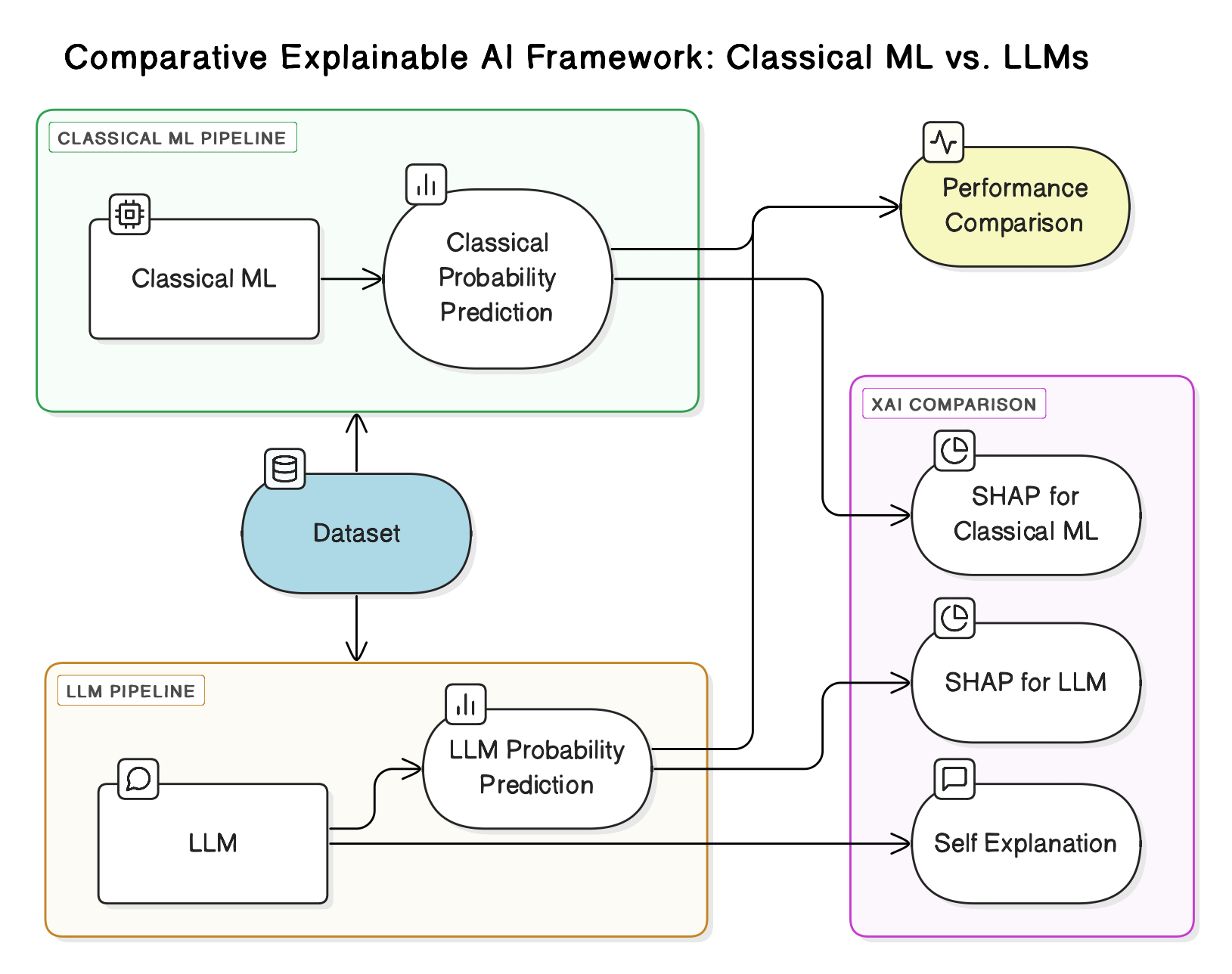}
    \caption{\small Comparative Explainable AI Framework: Classical ML vs. LLMs. The dataset is processed through two paradigms: (i) a structured LightGBM model and (ii) a zero-shot LLM using natural language prompts. Both generate probability predictions, analyzed individually and in an ensemble. Explainability is assessed via SHAP (for both) and LLM self-explanations, evaluating their alignment.}
    \Description{Diagram comparing explainable AI frameworks for classical machine learning and large language models, showing data flow and explanation methods.}
    \label{fig:Comparative}
\end{figure}

%%%%%%%%%%%%%%%%%%%%%%%%%%%%%%%%%%%%%%%%%%%%%%%%%%%%
\subsection{Inference Setup}

We systematically evaluate the predictive performance and explainability of LightGBM and zero-shot LLMs. We design the classification  problem such that both LightGBM and LLMs receive the same set of input features and output \textbf{probability estimates} instead of discrete classes. Probability outputs offer three advantages: (i) fine-grained evaluation via discrimination metrics (ROC-AUC, PR-AUC); (ii) enhanced explainability, as SHAP feature attribution is generally more informative when applied to probability scores rather than hard labels; and (iii) a direct test of LLMs’ capability as probability regressors.

%%%%%%%%%%%%%%%%%%%%%%%%%%%%%%%%%%%%%%%%%%%%%%%%%%%%
\subsection{Explainability}

We treat explainability as a model-auditing task focused on the faithfulness of the factors a model claims or appears to use. Our audit has two complementary components:

\begin{itemize}[leftmargin=*, rightmargin=0pt]
    \item \textbf{(A) SHAP as a model-agnostic audit baseline.} We use SHAP \cite{lundberg2017} to assign contribution values to each feature for both LightGBM and the LLM-based probabilistic classifier. SHAP values operationalize which features, and in what direction, are driving each model’s predictions.
    
    \item \textbf{(B) LLM self-explanations as unverified rationales.} In addition to SHAP, we prompt the LLM to provide instance-level rationales and feature-level directional\textbf{ }judgments (positive/negative/neutral). These self-explanations are treated as claims that must be checked against the SHAP audit checks.
\end{itemize}

We compare (i) global feature importance patterns (via SHAP) across models; (ii) directional dependence for key features (e.g., whether higher values increase or decrease repayment probability); and (iii) instance-level coherence between an LLM’s rationale and the corresponding SHAP attributions. Misalignment across these checks is interpreted as a faithfulness risk and a caution for deployment.

%%%%%%%%%%%%%%%%%%%%%%%%
\subsubsection{Scaling Shapley-Value Inference for LLMs}

To operationalize explainability as an \textit{audit} of model decision logic, we estimate feature attributions for both LightGBM and the zero-shot LLM classifiers using SHAP. Faithful auditing requires identifying which features actually drive predictions, not just producing plausible explanations.

We use SHAP for post hoc explanations \citep{lundberg2017}, specifically the model‑agnostic \texttt{\textbf{PermutationExplainer}}.
We adopt this efficient SHAP estimator because our prediction function is an LLM inference, which is costly. 
To balance accuracy and runtime, we sample 250 instances from each dataset for explanation. 
We construct the background (masker) via \(k\)-means clustering with \(C=5\) centroids (using \texttt{shap.kmeans}) and set the \texttt{max\_evals} budget so that the explainer executes exactly \(T=4\) permutations in our experiments.

\paragraph{\textbf{Approximate cost (model calls)}.}
Let \(K\) be the number of instances explained, \(M\) the number of features, \(B\) the number of background draws per masked evaluation (here \(B=C=5\)), and \(T\) the number of random permutations. The \textbf{PermutationExplainer} requires approximately:

\begin{equation}
\#\text{calls} \;\approx\; K \times T \times (M+1) \times B \;=\; \mathcal{O}(K\,T\,M\,B)
\end{equation}

model evaluations. In SHAP’s implementation, \(T\) is governed by \texttt{max\_evals} via the practical rule:

\begin{equation}
T \;\approx\; \Big\lfloor \frac{\texttt{max\_evals}}{2M} \Big\rfloor
\end{equation}

i.e., roughly \(2M\) masked evaluations per permutation path. With \(\texttt{max\_evals}=200\) and \(M=21\), this yields \(T=\lfloor 200/(2\times 21)\rfloor=4\). Using \(B=5\), the per‑instance cost is therefore \(\approx 4 \times (21{+}1) \times 5 = 440\) model calls.

\paragraph{\textbf{Why is it more efficient than} \texttt{KernelExplainer}?}
With a summarized background of \(C\) centroids, the dominant model‑call complexity of \texttt{KernelExplainer} scales as:

\begin{equation}
\#\text{calls} \;\approx\; K \times C \times M^2 \;=\; \mathcal{O}(K\,C\,M^2)
\end{equation}

due to sampling coalitions and fitting a kernel‑weighted regression. In our setting (\(C=5\), \(M=21\)) this is \(\approx 5 \times 21^2 = 2205\) evaluations per instance. By contrast, \texttt{PermutationExplainer} scales linearly in \(M\) and avoids the regression solve, yielding an expected per‑instance reduction of

\begin{equation}
\text{speedup} \;\approx\; \frac{C\,M^2}{T\,(M+1)\,B} \;\approx\; \frac{2205}{440} \;\approx\; 5\times
\end{equation}

The fivefold decrease in model calls translates into substantially lower LLM inference time while maintaining faithful attributions, which is why we use \texttt{PermutationExplainer} with \(T=4\).

%%%%%%%%%%%%%%%%%%%%%%%%
\subsubsection{LLM Self-Explanations}

Motivated by the emerging reasoning capabilities of large language models (LLMs) \cite{wei2022chain}, we use the LLM as an explainability tool alongside SHAP. Specifically, for each input feature, we provide its description to the LLM and prompt it to predict whether the feature is likely to have a positive, negative, or no effect on the predicted outcome. The LLM also generates a brief textual justification for each directional prediction, which we refer to as its self-explanation.

These LLM-generated self-explanations are treated as unverified rationales and are not assumed to reflect the model’s actual internal reasoning. We compare them against SHAP-based feature attributions, which serve as a model-agnostic baseline. When LLM explanations diverge from SHAP attributions, we interpret this misalignment as a potential risk of explainability, an indication that LLM may produce externally plausible but internally inconsistent reasoning. While we do not perform formal risk scoring, highlighting such discrepancies can inform governance decisions in high-stakes financial applications, where model transparency is essential.

Figure \ref{fig:Comparative}  illustrates the overall methodological framework. The dataset is processed through two parallel pipelines: (i) a structured LightGBM model and (ii) a zero-shot LLM using natural language prompts. Both produce probability estimates, which are then analyzed for predictive performance and explainability through SHAP and LLM self-explanations, enabling a cross-comparison of their decision logic.
\section{Experiments}

%%%%%%%%%%%%%%%%%%%%%%%%%%%%%%%%%%%%%%%%%%%%%%%%%%%%
\subsection{Data}

%%%%%%%%%%%%%%%%%%%%%%%%
\subsubsection{Task Description}

Loan default prediction is a critical challenge in credit risk assessment, where lenders estimate the likelihood of borrowers failing to repay their loans. Accurate predictions enable financial institutions to make informed lending decisions, set appropriate interest rates, and manage credit risk effectively. Because loan defaults directly affect credit access, financial stability, and regulatory compliance, this task is widely regarded as a high-stakes benchmark for testing the safety and reliability of predictive models in finance.

%%%%%%%%%%%%%%%%%%%%%%%%
\subsubsection{Dataset Description}

We use LendingClub’s publicly available loan records hosted by Kaggle \footnote{\url{https://www.kaggle.com/datasets/sndpred/loan-data}}, which were disclosed as part of the company’s regulatory filings with the U.S. Securities and Exchange Commission (SEC). As a major peer-to-peer lending platform, LendingClub was required to provide detailed loan issuance and performance data to comply with SEC regulations, making this dataset a widely used benchmark in credit risk modeling. The dataset includes loan applications issued between 2007 and 2016, with approximately 396{,}000 loan records. Each loan is labeled with its repayment status, distinguishing between \textit{Fully Paid} and \textit{Charged Off} (defaulted) loans. It contains both numerical and categorical attributes describing borrower creditworthiness and loan characteristics, for a total of 26 financial and credit-related features.

%%%%%%%%%%%%%%%%%%%%%%%%
\subsubsection{Preprocessing}

We removed five features due to high cardinality, redundancy, or potential data leakage:

\begin{itemize}[leftmargin=*, rightmargin=0pt]
    \item \texttt{issue\_d}: Loan issue date (introduces temporal bias)
    
    \item \texttt{earliest\_cr\_line}: Borrower’s earliest credit line (high cardinality)
    
    \item \texttt{address}: High-cardinality feature (introduces geographic bias)
    
    \item \texttt{emp\_title}: High-cardinality categorical variable
    
    \item \texttt{title}: High-cardinality; redundant with the \texttt{purpose} variable
\end{itemize}

We also excluded categorical features with more than 40 unique categories to mitigate overfitting in LightGBM. This ensures a fair comparison between models, preventing disadvantages for LightGBM (which lacks natural text processing) and restricting LLMs from exploiting external knowledge, such as macroeconomic trends from loan dates or geographic signals from addresses. These steps help control for potential bias and improve the fairness and auditability of the evaluation.

%%%%%%%%%%%%%%%%%%%%%%%%
\subsubsection{Final Dataset}

After preprocessing, the final dataset consists of 396{,}000 rows and 21 predictors (12 numerical and 9 categorical). The data was randomly split into training (80\%) and testing (20\%), with 79{,}200 instances used for LLM inference. This controlled setup ensures that both models operate on identical structured inputs without access to external priors, supporting a transparent and auditable comparison.

In Table \ref{tab:final_dataset_features}, we list all features. For numerical features, we report the interval bounded by the 1st and 99th percentiles. For categorical features, we report the values space (if it is not too large).

\begin{table}[t]
  \caption{Final Dataset Features.}
  \label{tab:final_dataset_features}
  \centering
  \begingroup
  \normalsize % or: \fontsize{10}{12}\selectfont
  \setlength{\tabcolsep}{4pt} % default ~6pt; smaller = tighter columns
  \renewcommand{\arraystretch}{0.9} % a touch more row height for readability

  \begin{tabularx}{\linewidth}{
      @{}>{\raggedright\arraybackslash}p{0.40\linewidth}
      @{\hspace{6pt}} % exact gap between the two columns
      >{\raggedright\arraybackslash}X@{}
  }
    \toprule
    
    \textbf{Feature Description} & \textbf{Range} \\
    
    \midrule
    
    \texttt{Loan amount} & [1600.0, 35000.0] \\
    \texttt{Term} & categorical: \{36 months, 60 months\} \\
    \texttt{Interest rate} & [6.03, 25.29] \\
    \texttt{Installment} & [55.32, 1204.57] \\
    \texttt{Grade} & categorical: \{A, B, C, D, E, F, G\} \\
    \texttt{Sub-grade} & categorical: \{A1, A2, …, G4, G5\} \\
    \texttt{Employment length} & categorical: \{<1 year, 1 year, …, 10+ years\} \\
    \texttt{Home ownership} & categorical: \{MORTGAGE, NONE, OTHER, OWN, RENT\} \\
    \texttt{Annual income} & [19000.0, 250000.0] \\
    \texttt{Verification status} & categorical: \{Not Verified, Source Verified, Verified\} \\
    \texttt{Purpose} & categorical: 14 values (e.g., wedding) \\
    \texttt{Debt-to-income (DTI) ratio} & [1.6, 36.41] \\
    \texttt{Open credit accounts} & [3.0, 27.0] \\
    \texttt{Public records} & [0.0, 2.0] \\
    \texttt{Revolving balance} & [169.05, 83505.9] \\
    \texttt{Revolving utilization rate} & [1.2, 98.0] \\
    \texttt{Total accounts} & [6.0, 60.0] \\
    \texttt{Initial listing status} & categorical: \{f, w\} \\
    \texttt{Application type} & categorical: \{DIRECT PAY, INDIVIDUAL, JOINT\} \\
    \texttt{Mortgage accounts} & [0.0, 9.0] \\
    \texttt{Public record bankruptcies} & [0.0, 1.0] \\
    
    \bottomrule
  \end{tabularx}
  \endgroup
\end{table}

%%%%%%%%%%%%%%%%%%%%%%%%%%%%%%%%%%%%%%%%%%%%%%%%%%%%
\subsection{Models}

%%%%%%%%%%%%%%%%%%%%%%%%
\subsubsection{LightGBM Training}

We trained a LightGBM classifier using the gradient boosting decision tree (GBDT) algorithm with a binary objective and AUC as the primary evaluation metric. The model was trained with a learning rate of 0.01 and up to 10,000 estimators, applying early stopping after 100 rounds based on validation performance. To reduce overfitting, we applied a feature fraction of 0.8, bagging fraction of 0.8, and L1/L2 regularization (0.1 each). The \texttt{num\_leaves} parameter was set to 63 and \texttt{min\_data\_in\_leaf} to 50. 
%No hyperparameter tuning was performed.

\begin{tcolorbox}[
  title=Instance-Level Prompt Template,
  colback=gray!10,
  colframe=gray!50,
  coltitle=black,
  colbacktitle=gray!15,
  fonttitle=\bfseries,
  boxrule=0.4pt,
  sharp corners=southwest,
  breakable,
  left=1mm,right=1mm,top=1mm,bottom=1mm
]
\small
Predict whether a loan application will be \texttt{"Fully Paid"} or \texttt{"Charged Off"} based on the borrower’s information. Use the features provided below to assess the likelihood of the loan being fully repaid.

\textbf{Loan Application Details:}

\begin{tcblisting}{listing only, colback=white, colframe=gray!40, boxrule=0.3pt, left=1mm,right=1mm,top=1mm,bottom=1mm}
<feature_1 name>: <feature_1 value>
...
<feature_N name>: <feature_N value>
\end{tcblisting}

Provide your estimated probability of the loan being \texttt{"Fully Paid"}. Also provide a brief explanation of this estimate based on the features. Your answer should only contain the probability estimate and the explanation in the following JSON format.

\begin{tcblisting}{listing only, colback=white, colframe=gray!40, boxrule=0.3pt, left=1mm,right=1mm,top=1mm,bottom=1mm}
{
  "Estimated Fully Paid Probability": <float value between 0 and 1>,
  "Explanation": <string value>
}
\end{tcblisting}
\end{tcolorbox}

\begin{tcolorbox}[
  title=Feature-Level Prompt Template,
  colback=gray!10,
  colframe=gray!50,
  coltitle=black,
  colbacktitle=gray!15,
  fonttitle=\bfseries,
  boxrule=0.4pt,
  sharp corners=southwest,
  breakable,
  left=1mm,right=1mm,top=1mm,bottom=1mm
]
\small
You are working on predicting whether a loan application will be "Fully Paid" or "Charged Off" based on the borrower’s information.
One of the features is the following:

\begin{tcblisting}{listing only, colback=white, colframe=gray!40, boxrule=0.3pt, left=1mm,right=1mm,top=1mm,bottom=1mm}
<feature name>
\end{tcblisting}

Do you think that this feature will impact the application positively, negatively, or have no impact?
Provide your answer as one of the three strings: \texttt{positive | negative | neutral}. Use the following JSON format:

\begin{tcblisting}{listing only, colback=white, colframe=gray!40, boxrule=0.3pt, left=1mm,right=1mm,top=1mm,bottom=1mm}
{
  "Feature impact": <positive | negative | neutral>
}
\end{tcblisting}
\end{tcolorbox}

We use LightGBM as a transparent and interpretable baseline, providing a benchmark for feature importance and prediction stability against which the behavior of large language models (LLMs) can be audited.

%%%%%%%%%%%%%%%%%%%%%%%%
\subsubsection{LLM Inference}

We evaluated three recent open-source instruction tuned LLMs of comparable size: \textbf{LLaMA-3.1-8B-Instruct} \cite{dubey2024llama}, \textbf{Gemma-2-9B-Instruct} \cite{team2024gemma}, and \textbf{Qwen-2.5-7B-Instruct} \cite{yang2024qwen2}. These models were pre-trained on 15.6T, 8T, and 18T tokens, respectively. All experiments used their instruction-tuned versions. Model weights were downloaded from the Hugging Face Hub \cite{wolf2020transformers}, and inference was run locally using vLLM\footnote{\url{https://github.com/vllm-project/vllm}} on four Nvidia A10G 24GB GPUs.

To reduce computation time, we randomly sampled 250 test instances for SHAP value estimation. We did not perform any fine-tuning and used strict zero-shot inference without any in-context learning. This design ensures the LLMs rely solely on the provided structured features, preventing data contamination or leakage from pre-training.

We used the same prompt templates for all LLMs. The \textbf{instance-level prompt} asked the model to jointly predict the probability that a loan would be fully repaid (a float between 0 and 1) and to provide a brief explanation. The \textbf{feature-level prompt} asked whether each feature would impact the prediction positively, negatively, or not at all.

\lstset{basicstyle=\ttfamily\small}

Unlike free-form textual descriptions, the structured dictionary format used for instance-level prompts is unlikely to have appeared in the LLMs’ pretraining corpus, which reduces the risk of data contamination.
\section{Analysis}

%%%%%%%%%%%%%%%%%%%%%%%%%%%%%%%%%%%%%%%%%%%%%%%%%%%%
\subsection{Performance Results}

\begin{figure}[t]
    \centering
    \includegraphics[width=1\linewidth]{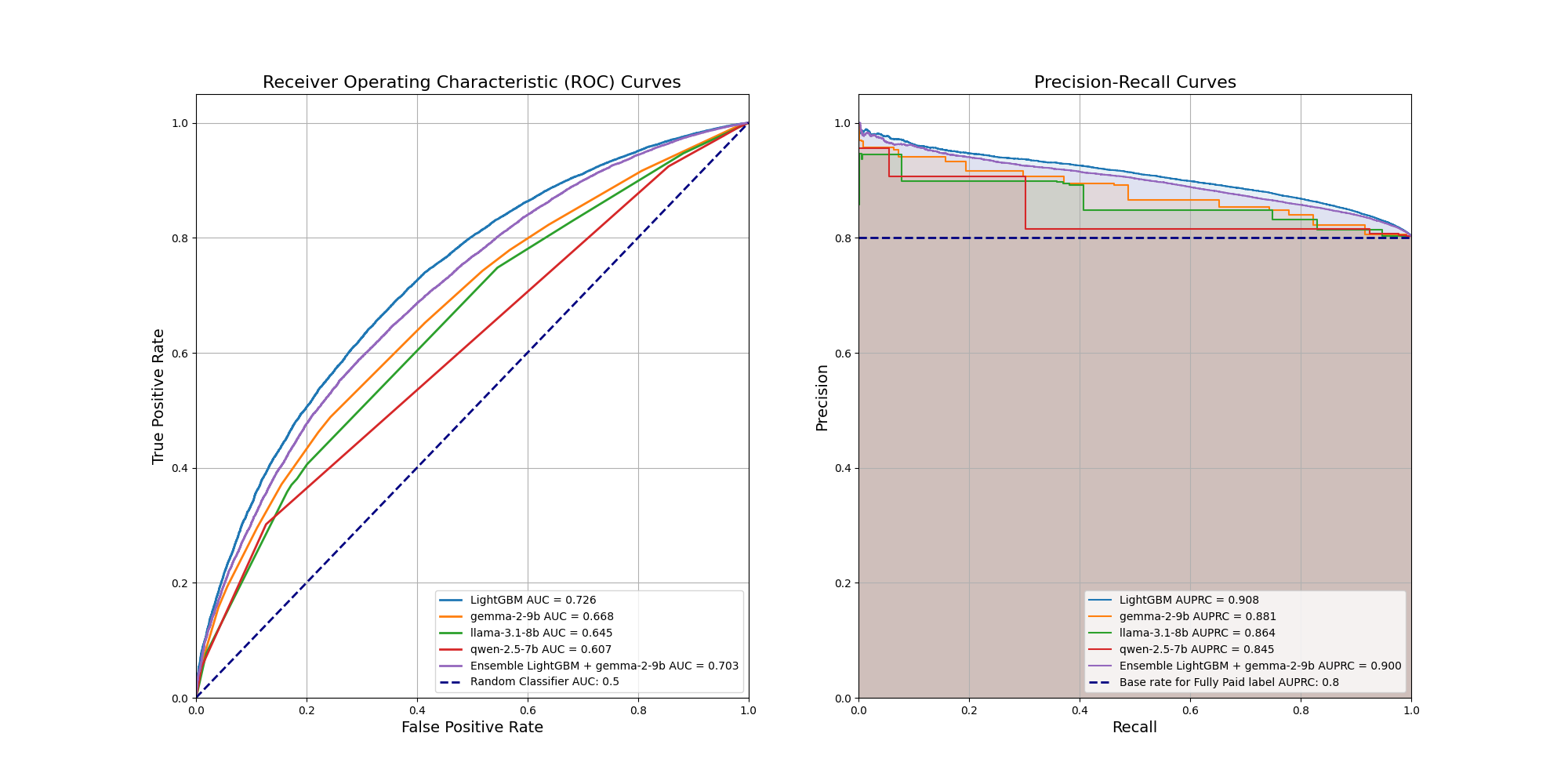}
    \caption{ROC and Precision-Recall curves comparing the performance of zero-shot LLMs and LightGBM on the loan classification task. LightGBM consistently outperforms individual LLMs with Gemma-2-9b showing the most promising result out of the LLMs.}
    \label{fig:ROC}
\end{figure}

Table \ref{tab:performance} and Figure \ref{fig:ROC} compare the performance of zero-shot LLMs and LightGBM on the loan repayment prediction task. LightGBM achieved the highest ROC–AUC (0.73), outperforming all LLMs in the zero-shot setting. Among LLMs, Gemma-2-9B performed best (0.67), followed by LLaMA-3.1-8B (0.65) and Qwen-2.5-7B (0.61). These findings are consistent with prior evidence that gradient boosting often surpasses deep learning methods on structured tabular data. The LightGBM–Gemma-2-9B ensemble achieved 0.70 ROC–AUC, indicating no diversification benefit over LightGBM alone.

PR–AUC results follow a similar pattern. LightGBM obtained the highest PR–AUC (0.91), exceeding all LLMs. Gemma-2-9B again ranked highest among the LLMs (0.88), followed by LLaMA-3.1-8B (0.86) and Qwen-2.5-7B (0.85). The ensemble model reached 0.90, closely trailing LightGBM. All models outperformed the base-rate PR–AUC (0.80), confirming meaningful predictive signal.

Overall, these results show that while zero-shot LLMs achieve reasonable predictive performance, they remain inferior to a well-tuned LightGBM model on structured financial data, underscoring the need for careful governance if deployed in high-stakes settings.

\begin{table}[t]
\begingroup
\setlength{\abovecaptionskip}{4pt}   
\setlength{\belowcaptionskip}{2pt}   
\setlength{\tabcolsep}{4pt}          
\centering
\caption{Performance Metrics of Models for Loan Default Classification}
\label{tab:performance}

% keep caption normal size; only the grid a bit smaller if you like:
{\footnotesize
\fontsize{8}{10}\selectfont
\begin{tabular}{p{0.6\linewidth}cc}
\toprule
\textbf{Model} & \textbf{ROC-AUC} & \textbf{PR-AUC} \\
\midrule
LightGBM & \textbf{0.73} & \textbf{0.91} \\
Gemma-2-9B & 0.67 & 0.88 \\
Llama-3.1-8B & 0.65 & 0.86 \\
Qwen-2.5-7B & 0.61 & 0.85 \\
Ensemble (LightGBM + Gemma-2-9B) & 0.70 & 0.90 \\
\midrule
Random Classifier (Baseline) & 0.50 & -- \\
Base Rate for Fully Paid Label (Baseline) & -- & 0.80 \\
\bottomrule
\end{tabular}
}
\endgroup
\end{table}

% tiny nudge so the heading hugs the table
\vspace{-0.35\baselineskip}

\subsection{SHAP Feature Importance Comparison}
\begin{figure}
    \centering
    \includegraphics[width=0.3\textwidth]{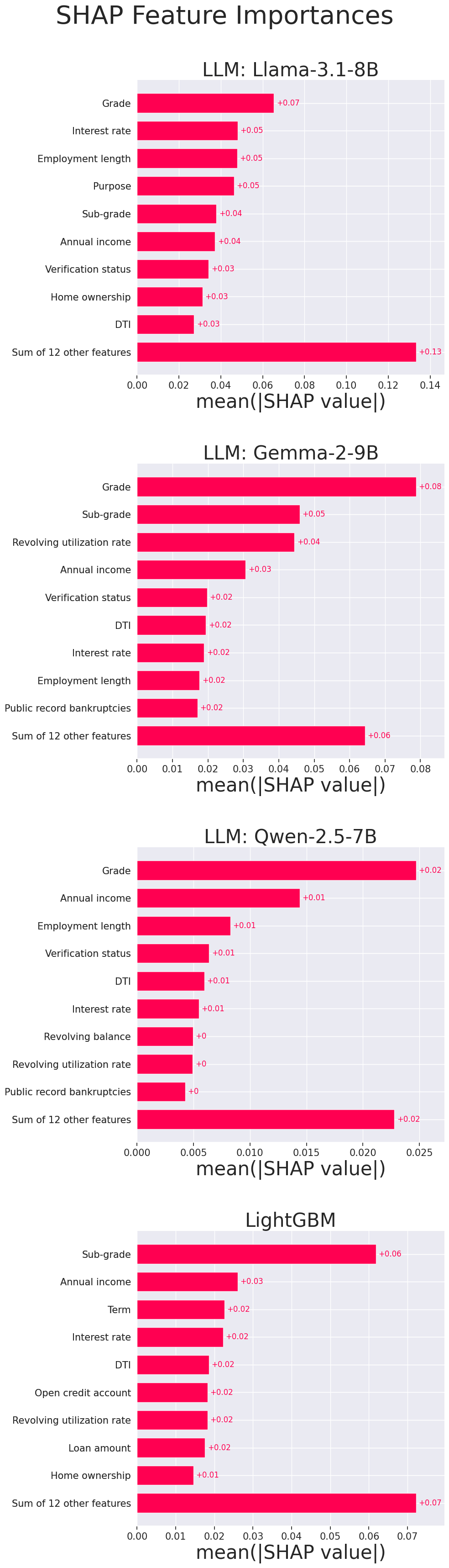}
    \caption{SHAP feature importance comparison between LightGBM and LLMs. Despite being in a zero-shot setting, LLMs identify a remarkably similar set of key financial features as LightGBM, though with differences in feature weighting and distribution.}
    \label{fig:SHAP}
\end{figure}

Figure \ref{fig:SHAP} presents the SHAP feature importance rankings for LightGBM and the three LLMs, providing insight into the key factors influencing loan classification decisions across models. The corresponding numerical values are reported in Table \ref{tab:feature_importance}.

A primary observation is the strong overlap in the top-ranked features across all models, despite the LLMs operating in a zero-shot setting. Features such as \texttt{Sub-grade}, \texttt{Grade}, \texttt{Annual income}, and \texttt{Interest rate} consistently emerge as dominant predictors. This suggests that LLMs, even without task-specific fine-tuning, are able to extract and prioritize meaningful financial attributes in a manner broadly consistent with classical ML models like LightGBM.

However, notable differences emerge in feature weighting and rank order. LightGBM assigns greater relative importance to structured numerical features such as \texttt{Sub-grade}, \texttt{Annual income}, and \texttt{Loan amount}, reflecting its reliance on directly interpretable numerical signals. In contrast, the LLMs particularly LLaMA-3.1-8B and Gemma-2-9B distribute their importance more evenly across categorical and behavioral attributes such as \texttt{Verification status}, \texttt{Purpose}, and \texttt{Home ownership}. This indicates that LLMs may be leveraging latent semantic relationships within categorical features that are not explicitly modeled by LightGBM.

\begin{table}[t]
\centering
\caption{Feature Importance Comparison}
\label{tab:feature_importance}
\begin{tabular}{p{0.3\linewidth}
                >{\centering\arraybackslash}p{0.12\linewidth}
                >{\centering\arraybackslash}p{0.12\linewidth}
                >{\centering\arraybackslash}p{0.12\linewidth}
                >{\centering\arraybackslash}p{0.12\linewidth}}
                
\toprule

\textbf{Feature} & \textbf{LGBM} & \textbf{Gemma-2-9B} & \textbf{Llama-3.1-8B} & \textbf{Qwen-2.5-7B} \\

\midrule

\texttt{Sub-grade}         & 0.062 & 0.046 & 0.038 & 0.004 \\
\texttt{Annual income}     & 0.026 & 0.031 & 0.037 & 0.014 \\
\texttt{Term}              & 0.023 & 0.002 & 0.008 & 0.000 \\
\texttt{Interest rate}     & 0.022 & 0.019 & 0.048 & 0.005 \\
\texttt{DTI}               & 0.019 & 0.019 & 0.027 & 0.006 \\
\texttt{Open account}      & 0.018 & 0.004 & 0.007 & 0.001 \\
\texttt{Revolving util}    & 0.018 & 0.044 & 0.021 & 0.005 \\
\texttt{Loan amount}       & 0.018 & 0.006 & 0.026 & 0.004 \\
\texttt{Home ownership}    & 0.015 & 0.012 & 0.031 & 0.003 \\
\texttt{Grade}             & 0.013 & 0.079 & 0.065 & 0.025 \\

\bottomrule

\end{tabular}
\end{table}

%%%%%%%%%%%%%%%%%%%%%%%%%%%%%%%%%%%%%%%%%%%%%%%%%%%%
\subsection{SHAP Comparative Analysis}
\label{sec:shap_comparative}

\begin{figure}[t]
    \centering
    \includegraphics[width=1\linewidth]{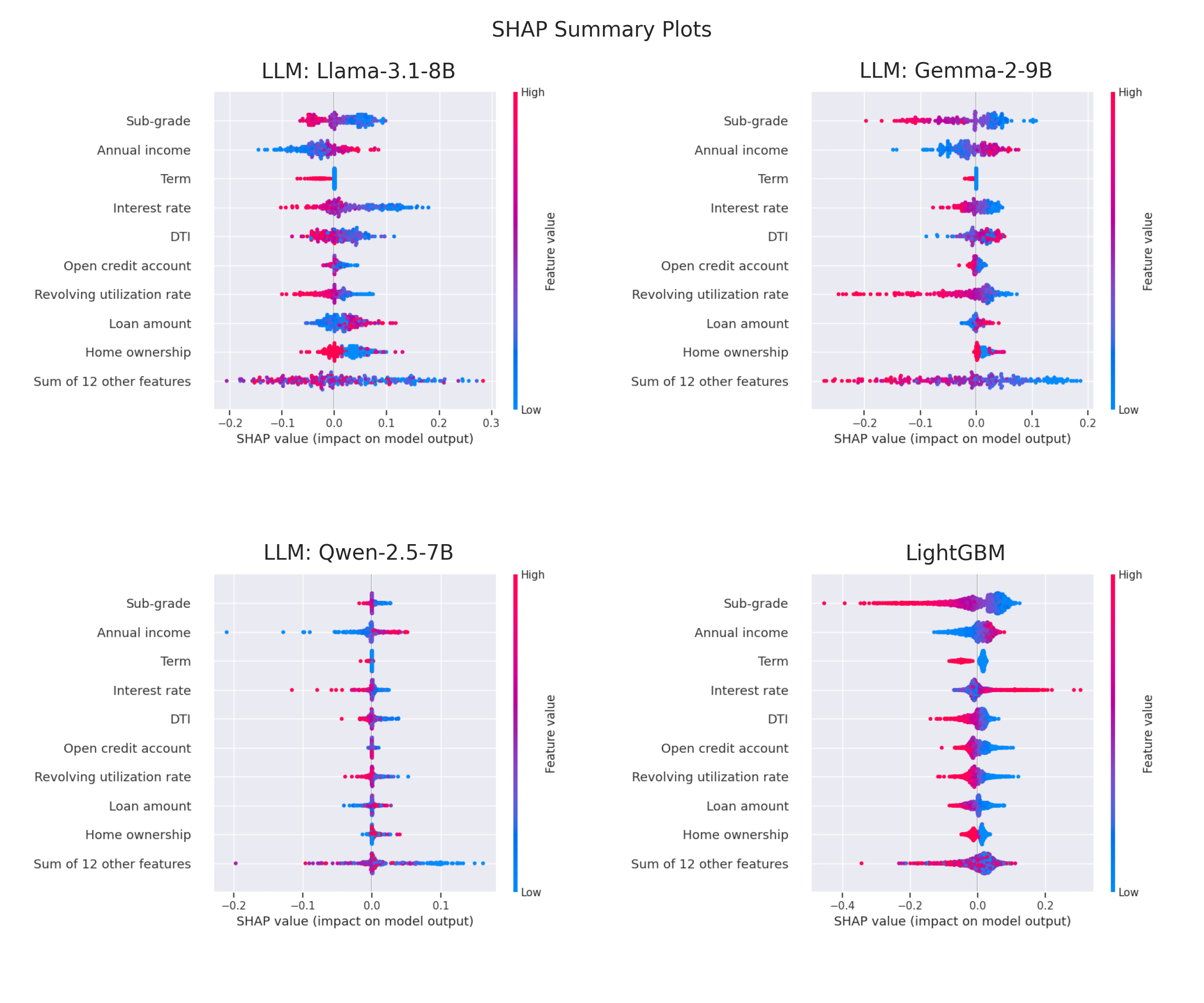}
    \vspace{-3em}
    \caption{\small SHAP summary plots comparing feature importance distributions for LLMs and LightGBM. LightGBM shows a more structured reliance on key financial indicators, while LLMs exhibit more dispersed and lower-magnitude SHAP values, indicating weaker feature dependencies.}
    \label{fig:SHAPsummary}
\end{figure}

To further investigate the decision mechanisms of the models, we compare the SHAP summary plots of LightGBM and the three LLMs. In Figure \ref{fig:SHAPsummary}, these plots illustrate how variations in feature values influence predicted loan repayment probabilities.

Despite operating in a zero-shot setting, the LLMs successfully extract meaningful relationships from structured financial features. Core predictors such as \texttt{Sub-grade}, \texttt{Interest rate}, and \texttt{Loan amount} consistently emerge as important across all models, suggesting that the LLMs are able to capture many of the same risk-relevant factors identified by LightGBM. Notably, Gemma-2-9B assigns comparatively higher SHAP values to these features, aligning with its superior classification performance among the evaluated LLMs.

While the models converge on key features, their attribution patterns also reveal important differences. LightGBM places substantial weight on well-established numerical predictors such as \texttt{Sub-grade} and \texttt{Annual income}, whereas the LLMs distribute importance more broadly across behavioral and categorical variables. This more diffuse attribution pattern implies that the LLMs may be leveraging latent feature interactions rather than relying solely on direct numerical signals. Such behavior could reflect their ability to encode semantic relationships across variables that classical models do not capture explicitly.

A further notable observation is the reversal of SHAP relationships for certain features between the models. For example, in LightGBM, higher \texttt{Interest rate} values are associated with a greater predicted probability of being fully repaid (positive SHAP impact), whereas all three LLMs display the opposite trend. Similarly, \texttt{DTI} exhibits a different effect in Gemma-2-9B compared to the other LLMs and LightGBM. These discrepancies suggest that while the LLMs extract informative patterns, they may rely on internally learned representations that diverge from classical feature logic, which has implications for their reliability in regulated financial contexts.

\begin{figure}[t]
    \centering
    \includegraphics[width=1\columnwidth]{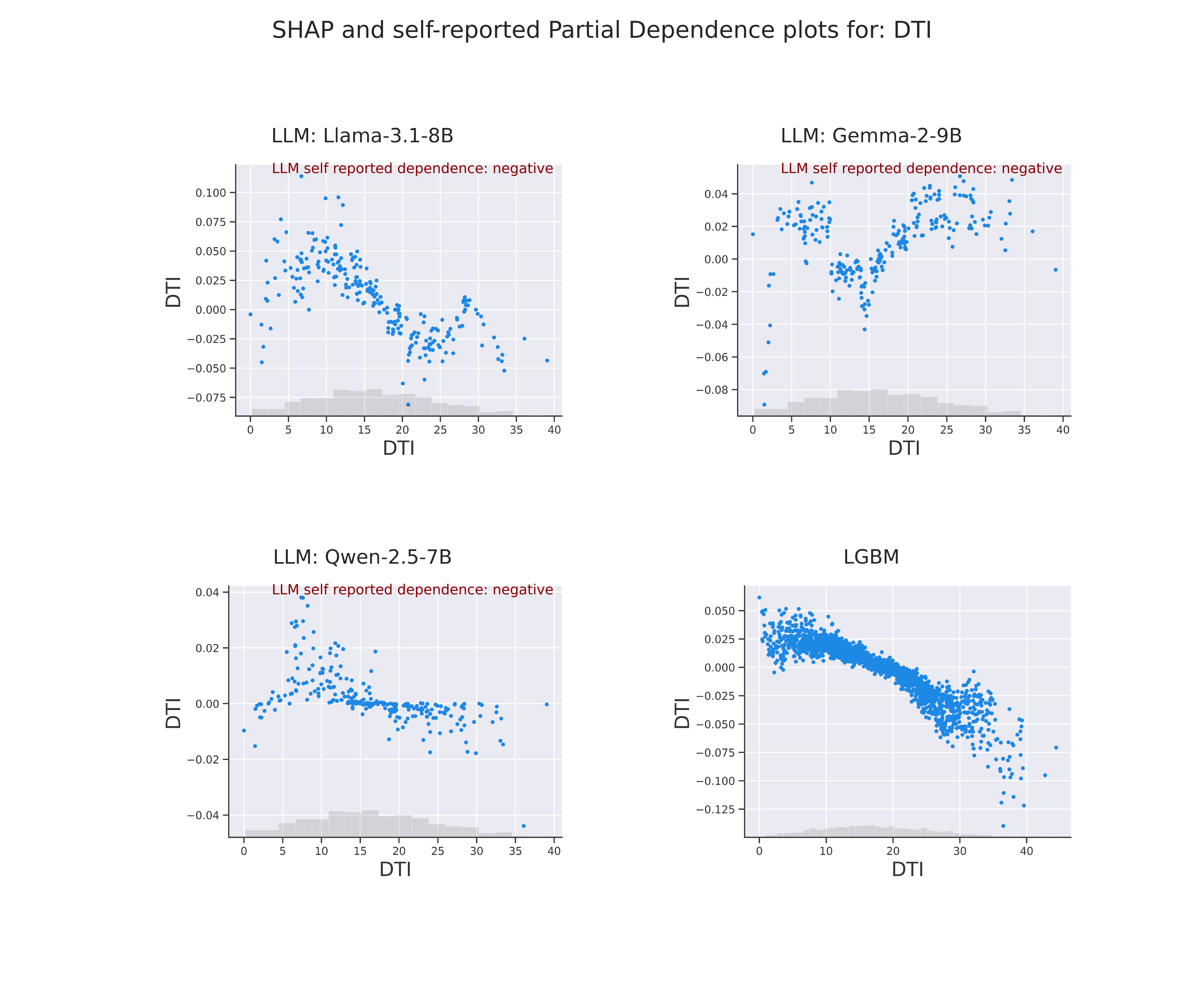}
    \vspace{-3em}
    \caption{SHAP Feature dependence plots and LLM self-explanations for the feature \texttt{DTI}.}
    \label{fig:DTI}
\end{figure}

\begin{figure}[t]
    \centering
    \includegraphics[width=1\columnwidth]{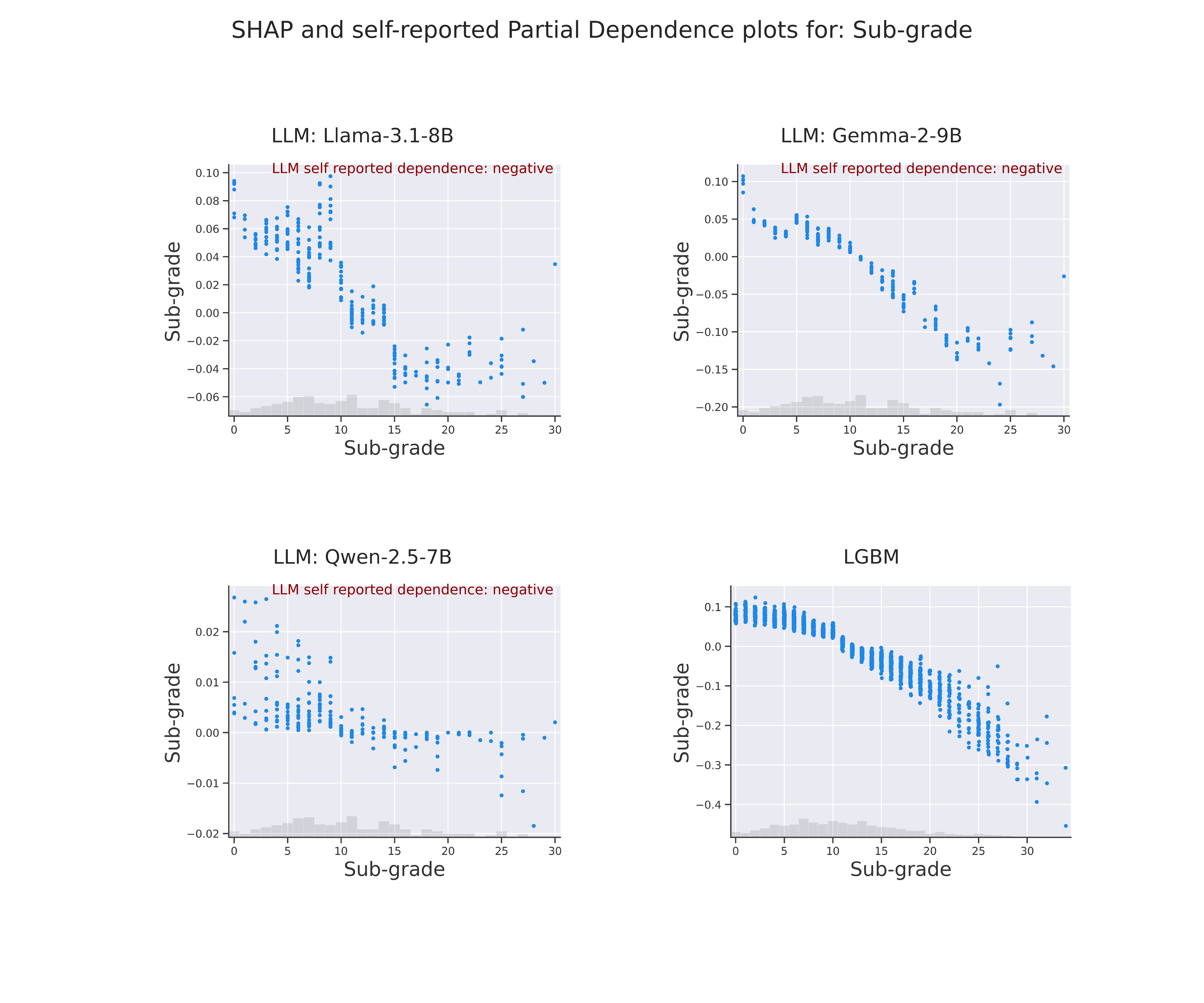}
    \vspace{-3em}
    \caption{SHAP Feature dependence plots and LLM self-explanations for the feature \texttt{Sub-grade}.}
    \label{fig:ReportedDependence}
\end{figure}

%%%%%%%%%%%%%%%%%%%%%%%%%%%%%%%%%%%%%%%%%%%%%%%%%%%%
\subsection{SHAP Feature Dependences and LLM Self-Explanation}
\label{sec:shap_vs_self}

We compare classical model-centric explainability (SHAP dependence plots) with LLM self-reported explanations for two key features: \texttt{DTI} (Debt-to-Income ratio) and \texttt{Sub-grade}. LLM self-reported dependence is obtained by directly prompting the models on whether each feature exerts a positive, negative, or neutral effect on loan repayment likelihood, and is shown at the top of each chart.  

\paragraph{\textbf{\texttt{DTI} (Debt-to-Income ratio)}} All three LLMs self-report a negative relationship between \texttt{DTI} and loan repayment likelihood, aligning with their SHAP dependence plots (Figure~\ref{fig:DTI}). LightGBM also shows a clear negative dependence. However, Gemma-2-9B displays a notable inconsistency: its SHAP values suggest a positive contribution at higher \texttt{DTI} levels, contradicting its own self-explanation.

\paragraph{\textbf{\texttt{Sub-grade}}} Both the LLM SHAP plots and self-reports consistently show \texttt{Sub-grade} as a feature negatively correlated with default risk (A1 being lowest risk and G5 highest) (Figure \ref{fig:ReportedDependence}). LightGBM exhibits a strong and nearly monotonic negative dependence, indicating heavy reliance on \texttt{Sub-grade} for risk discrimination. LLMs reproduce this general trend but with shallower slopes and greater local variability.  

\paragraph{\textbf{Summary of alignment}} While LLM self-explanations often align with their SHAP dependence patterns, there are notable divergences such as \texttt{DTI}, which was self-reported as negative but showed positive SHAP impact at higher values. Such mismatches suggest that LLM self-explanations, while often plausible, do not always reflect their internal decision-making mechanisms. Unlike LightGBM, which captures purely statistical relationships from the structured dataset, LLMs may incorporate latent financial priors beyond the data. These inconsistencies reinforce the need for independent audits before trusting LLM self-explanations in high-stakes financial workflows.
\section{Conclusion}
\label{sec:conclusion}

The growing adoption of large language models (LLMs) for structured classification raises critical questions about their validity and safety in high-stakes financial decision-making. This study systematically compared zero-shot LLM classifiers with LightGBM on a structured loan default prediction task, evaluating both predictive performance and explainability through SHAP-based audits.  

Our findings reveal that while LLMs can capture several key statistical patterns similar to LightGBM, they remain inferior in predictive accuracy, with LightGBM achieving the highest ROC--AUC and PR--AUC scores. LLM-generated self-explanations occasionally align with SHAP feature attributions, but observed discrepancies indicate that these rationales may rely on external priors rather than purely dataset-driven reasoning. Moreover, ensembling LLMs with LightGBM did not yield meaningful performance gains, suggesting limited complementarity between the two paradigms.  

These results highlight  that while LightGBM remains the more reliable choice when skilled data scientists are available, LLMs could serve as a practical fallback in small-data settings where fine-tuning is infeasible, provided their outputs are independently audited. Future research should explore fine-tuned LLMs for tabular modeling, hybrid approaches that better integrate structured and unstructured reasoning, and establish robust reliability and fairness assessments to ensure their responsible deployment in financial applications.

\bibliographystyle{ACM-Reference-Format}
\bibliography{sample-base}

\end{document}